\documentclass[acmtog,authorversion]{acmart}

\usepackage[ruled]{algorithm2e} 
\usepackage{pifont} 
\usepackage{multirow}
\usepackage{subcaption}

\SetAlFnt{\small}
\SetAlCapFnt{\small}
\SetAlCapNameFnt{\small}
\SetAlCapHSkip{0pt}

\acmJournal{TOG}

\copyrightyear{2025}
\acmYear{2025}
\acmConference[SA Conference Papers '25]{SIGGRAPH Asia 2025 Conference Papers}{December 15--18, 2025}{Hong Kong, Hong Kong}
\acmBooktitle{SIGGRAPH Asia 2025 Conference Papers (SA Conference Papers '25), December 15--18, 2025, Hong Kong, Hong Kong}\acmDOI{10.1145/3757377.3763870}
\acmISBN{979-8-4007-2137-3/2025/12}

\acmDOI{10.1145/3757377.3763870}

\begin{document}
\title{Teamwork: Collaborative Diffusion with Low-rank Coordination
  and Adaptation}

\author{Sam Sartor}
\affiliation{%
  \institution{College of William \& Mary}
  \city{Williamsburg}
  \country{USA}
}
\orcid{0009-0001-1915-6887}
\email{slsartor@wm.edu}

\author{Pieter Peers}
\affiliation{%
  \institution{College of William \& Mary}
  \city{Williamsburg}
  \country{USA}
}
\orcid{0000-0001-7621-9808}
\email{ppeers@siggraph.org}

\renewcommand\shortauthors{Sartor and Peers}

\def\equationautorefname~#1\null{Equation~(#1)\null}
\def\sectionautorefname{Section}
\def\subsectionautorefname{Section}

\def\etal{{et al.}}
\def\ie{{i.e.}}
\def\eg{{e.g.}}

\def\BF{\textbf}
\def\UL{\underline}

\def\cmark{\ding{51}}%
\def\xmark{\ding{55}}%
\def\TODOM{\textcolor{red}{TODO}}

\newcount\numcols
\numcols=3 
\def\fraclinewidth{1.0}
\def\pad{0.03in}

\def\figureprefix{compressed_figures/}
\newlength{\colwidth}
\newlength{\colpad}
\newcommand{\setcolwidth}{%
  \setlength{\colwidth}{\dimexpr\fraclinewidth\linewidth/\the\numcols-\colpad+\colpad/\the\numcols-0.001pt\relax}%
  \setlength{\fboxsep}{0pt}%
  \setlength{\lineskip}{0pt}%
}

\newcommand{\anysquare}[1]{%
  \setcolwidth%
  \ifhmode\hspace{\colpad}\fi\parbox[c][\colwidth]{\colwidth}{#1}\ignorespaces%
}

\newcommand{\anycol}[1]{%
  \setcolwidth%
  \ifhmode\hspace{\colpad}\fi\parbox{\colwidth}{#1}\ignorespaces%
}

\newcommand{\txtcol}[1]{%
  \anycol{\centering #1}\ignorespaces%
}

\newcommand{\emptycol}{%
  \setcolwidth%
  \ifhmode\hspace{\colpad}\fi\hspace*{\colwidth}\ignorespaces%
}

\newcommand{\todosquare}{%
  \setlength{\fboxrule}{1pt}%
  \anysquare{\fbox{\textcolor{blue!40}{\rule{\dimexpr\colwidth-2pt}{\dimexpr\colwidth-2pt}}}}%
}

\newcommand{\qmarksquare}{%
  \setcolwidth%
  \setlength{\fboxsep}{0pt}%
  \setlength{\fboxrule}{1pt}%
  \fbox{%
    \begin{minipage}[c][\dimexpr\colwidth-2pt][c]{\dimexpr\colwidth-2pt}
      \centering
      \resizebox{!}{0.5\colwidth}{\textbf{?}}%
    \end{minipage}%
  }\ignorespaces%
}

\newcommand{\xmarksquare}{%
  \setcolwidth%
  \setlength{\fboxsep}{0pt}%
  \setlength{\fboxrule}{1pt}%
  \fbox{%
    \begin{minipage}[c][\dimexpr\colwidth-2.01pt][c]{\dimexpr\colwidth-2.01pt}
      \centering
      \resizebox{!}{0.5\colwidth}{\xmark}%
    \end{minipage}%
  }\ignorespaces%
}

\newcommand{\imgsquare}[1]{%
  \anysquare{\includegraphics[width=\colwidth]{\figureprefix#1}}\ignorespaces%
}

\newcommand{\fimgsquare}[1]{%
  \setlength{\fboxrule}{1pt}%
  \anysquare{\color{orange}\fbox{\normalcolor\includegraphics[width=\dimexpr\colwidth-2pt]{figures/#1}}}\ignorespaces%
}

\newcommand{\setcols}[2]{%
  \numcols=#1%
  \setlength{\colpad}{0pt}\relax%
  \setcolwidth%
  #2%
}
\newcommand{\setpadcols}[3]{%
  \numcols=#1%
  \setlength{\colpad}{#2}\relax%
  \setcolwidth%
  #3%
}

\newlength{\dumbfigurepad}
\setlength{\dumbfigurepad}{5pt}
\newcommand{\dumbfigure}[2]{%
  \setlength{\fboxsep}{\dumbfigurepad}%
  \setlength{\fboxrule}{0pt}%
  \fbox{\begin{minipage}{\dimexpr#1\textwidth-\dumbfigurepad-\dumbfigurepad}#2\end{minipage}}%
}

\definecolor{dblue}{rgb}{0.0,0.0,0.5}
\definecolor{dgreen}{rgb}{0.0,0.5,0.0}
\definecolor{dred}{rgb}{0.6,0.0,0.0}
\definecolor{dorange}{rgb}{0.6,0.25,0.0}
\definecolor{dyellow}{rgb}{0.5,0.5,0.0}
\newcommand{\NOTE}[1]{ \texttt{ \textbf{NOTE:} #1~ } }
\newcommand{\TODO}[1]{ \textcolor{red}{\texttt{ \textbf{TODO:} #1~ }} }
\newcommand{\CHECK}[1]{ \textcolor{purple}{\textbf{(CHECK)} #1~ }}
\newcommand{\PP}[1]{ \textcolor{blue}{\texttt{ \textbf{PP:} #1~ }} }
\newcommand{\SLS}[1]{ \textcolor{dgreen}{\texttt{ \textbf{SS:} #1~ }} }

\newcommand{\ignorethis}[1]{}

\providecommand{\DIFdel}[1]{}
\renewcommand{\DIFdel}[1]{}

\providecommand{\DIFadd}[1]{}
\renewcommand{\DIFadd}[1]{\textcolor{blue}{#1}}

\newcommand{\scheme}[1]{\textsc{#1}}
\newcommand{\prompt}[1]{\textsc{#1}}

\newcommand{\mtx}[1]{\mathbf{#1}}
\newcommand{\vect}[1]{\mathbf{#1}}
\newcommand{\axis}[1]{#1}

\def\W{\mtx{W}}
\def\A{\mtx{A}}
\def\B{\mtx{B}}
\def\T{T}
\def\R{\mathbb{R}}
\def\X{\vect{x}}
\def\H{\vect{y}}
\def\M{m}
\def\N{n}

\begin{abstract}

  Large pretrained diffusion models can provide strong priors
  beneficial for many graphics applications.  However, generative
  applications such as neural rendering and inverse methods such as
  SVBRDF estimation and intrinsic image decomposition require
  additional input or output channels. Current solutions for channel
  expansion are often application specific and these solutions can be
  difficult to adapt to different diffusion models or new tasks.  This
  paper introduces Teamwork: a flexible and efficient unified solution
  for jointly increasing the number of input and output channels as
  well as adapting a pretrained diffusion model to new
  tasks. Teamwork achieves channel expansion without altering the
  pretrained diffusion model architecture by coordinating and adapting
  multiple instances of the base diffusion model (\ie, teammates).  We
  employ a novel variation of Low Rank-Adaptation (LoRA) to jointly
  address both adaptation and coordination between the different
  teammates. Furthermore Teamwork supports dynamic (de)activation of
  teammates.  We demonstrate the flexibility and efficiency of
  Teamwork on a variety of generative and inverse graphics tasks such
  as inpainting, single image SVBRDF estimation, intrinsic
  decomposition, neural shading, and intrinsic image synthesis.
  \end{abstract}

\ccsdesc[500]{Computing methodologies~Image manipulation}

\keywords{Diffusion, Input-output Expansion, Collaborative Diffusion,
  LoRA, Inpainting, SVBRDF, Intrinsic Decomposition}

\maketitle

\begin{figure*}[ht!]
  \centering
  \setlength{\tabcolsep}{0.5cm}
  \begin{tabular}{ccccc}
    \small
    (a) Head Expansion &(b) ControlNet &(c) Batching &(d) Joint Attention &(e) Teamwork
    \vspace{1ex} \\
    \includegraphics[height=1.5in]{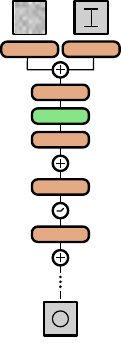} &
    \includegraphics[height=1.5in]{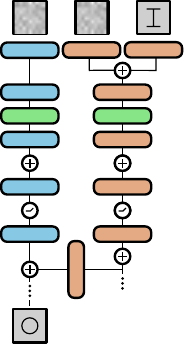} &
    \includegraphics[height=1.5in]{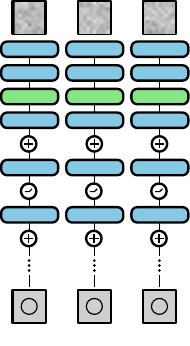} &
    \includegraphics[height=1.5in]{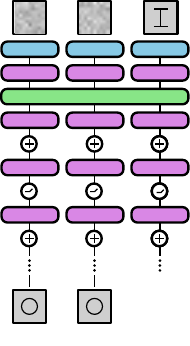} &
    \includegraphics[height=1.5in]{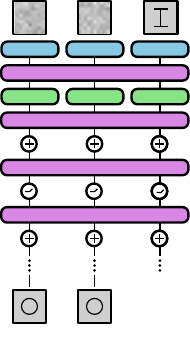}
  \end{tabular}
  \\ \vspace{1ex}
  \setcols{7}{
    \small
    \txtcol{\includegraphics[width=1em]{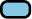} Frozen Linear}
    \txtcol{\includegraphics[width=1em]{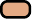} Trainable Linear}
    \txtcol{\includegraphics[width=1em]{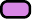} LoRA Linear}
    \txtcol{\includegraphics[width=1em]{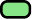} Attention}
    \txtcol{\includegraphics[width=1em]{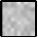} Noisy Latents}
    \txtcol{\includegraphics[width=1em]{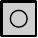} Output Latents}
    \txtcol{\includegraphics[width=1em]{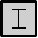} Input Latents}
  }
  \caption{Schematic overview of different common input and output
    channel expansion techniques for pretrained diffusion
    models. Input expansion: (a) Zero-convolution expands the input
    head with zero-initialized weights and subsequently finetunes the
    model. The latter operation increases the risk of overfitting, and
    thus destroy potentially valuable embedded priors. (b) ControlNet
    reduces the risk of overfitting by finetuning a copy of the
    (frozen) pretrained diffusion model while injecting weight offsets
    to each layer in the original diffusion model. Output expansion:
    (c) Batching, a common training optimization step, allows to run
    multiple instances of a model in parallel at inference. However,
    each model instance in a batch is unaware of the others, and thus
    no coordination occurs between the different instances. (d) Joint
    Attention coordinates between different instances of adaptations
    of a diffusion model by replacing self-attention layers with
    joint-attention layers over the multiple instances.  Conceptually,
    Joint Attention is the dual of Teamwork (e) which keeps attention
    computations within each instance, but instead shares the features
    from the linear layers.}
  \label{fig:archs}
\end{figure*}

\section{Introduction}
\label{sec:intro}

Diffusion models~\cite{Song:2021:SBG,Karras:2022:EDM,Rombach:2022:HRI}
are a versatile class of generative models that have been applied to a
wide range of image synthesis tasks such as image
restoration~\cite{Dhariwal:2021:DMB,Ho:2020:DDP,Ho:2022:CDM},
super-resolution~\cite{Kadhodaie:2021:SSL,Saharia:2023:ISR},
image-to-image translation~\cite{Sasaki:2021:UDU,Saharia:2022:PTI},
and text-to-image synthesis~\cite{Nichol:2022:GTP,Ramesh:2022:HTC}.
While flexible and powerful, diffusion models are also incredibly data
hungry and computationally expensive to train.  Consequently, recent
work leverages large pretrained text-to-image diffusion
models~\cite{StableDiffusion,DeepFloyd,VonPlaten:2022:Diffusers} as a
prior for different tasks.

Pretrained text-to-image diffusion models typically take, besides a
text prompt, a three-channel input (\ie, noisy image) and output an
three-channel RGB image, possibly via a latent space of different
dimensions. Many graphics applications, however, require additional
input channels (\eg, neural
shading~\cite{Nalbach:2017:DSC,Zeng:2024:RGB}) and/or output more than
three data channels (\eg, spatially-varying reflectance
distribution function (SVBRDF)
estimation~\cite{Sartor:2023:MGD,Vecchio:2023:CAC}, intrinsic image
decomposition~\cite{Zeng:2024:RGB,Kocsis:2023:IID}, and
matting~\cite{Zhang:2024:TIL}).  Common solutions for input expansion
include ControlNet~\cite{Zhang:2023:ACC} and adding zero-convolutions
to the head (\eg,~\cite{Sartor:2023:MGD,Zeng:2024:RGB}). Expanding the
number of output channels is less standardized and relies on bespoke
solutions ranging from shared/joint
attention~\cite{Hertz:2023:SAI,Zhang:2024:TIL} or adding specialized
tokens (\ie, prompt keywords) to switch between output tasks/channel
(\eg,~\cite{Zeng:2024:RGB,Luo:2024:IDJ}).  Expanding both input and
output channels requires a combination of disparate methods, which
each can differ how much embedded priors in the pretrained model are
preserved, how well the method mitigates overfitting, and training
costs.

In this paper, we present a flexible unified solution, named
\emph{Teamwork}, for expanding the number of input and output channels
when adapting a pretrained diffusion model. We formulate channel
expansion as adapting multiple instances of the same foundational base
model that each take care of three input or output channels.  In order
to model correlations between different channels, the different
adapted instances of the base model require a mechanism to coordinate
and exchange information.  Teamwork solves both problems, adaptation
and coordination, simultaneously via an elegant extension of
LoRA~\cite{Hu:2021:LRA} that models offsets to the linear layers over
\emph{all} base model instances jointly with an low-rank
approximation.  Teamwork offers a number of advantages over prior
work.  First, both input and output expansion is handled in the same
framework.  Second, Teamwork is easy to implement on top of a regular
LoRA implementation.  Third, Teamwork is efficient; it does not incur
any computation or memory overhead compared to adapting each
base-diffusion model separately with regular LoRA.  Finally, Teamwork
supports dynamic (de)activation of input and output channels without
retraining or finetuning. Not only does this offer additional
flexibility during inference, it also allows us, similar to
RGB$\rightarrow$X~\cite{Zeng:2024:RGB}, to train Teamwork with
heterogeneous datasets that feature different subsets of data channels
thereby facilitating enrichment and diversification of training
exemplars.

We demonstrate the efficacy and flexibility of our Teamwork framework
on a variety of inverse computer graphics tasks, such as single image
SVBRDF estimation and intrinsic image decomposition, as well as
generative computer graphics tasks, such as inpainting, neural
rendering, and intrinsic image synthesis.  Using these applications,
we validate Teamwork's performance against competing expansion and
coordination techniques, demonstrate the importance of coordination,
and explore the impact of dynamic (de)activation of input and output
channels.  Code and trained models can be found at
\url{https://github.com/samsartor/teamwork}.

\section{Related Work}
\label{sec:related}

\paragraph{Diffusion Models}
Diffusion models form a class of powerful generative machine learning
solutions that have the uncanny ability to synthesize high-quality
images covering a wide variety of styles ranging from photorealistic
to artistic.  However, image diffusion models are also incredibly data
hungry and computationally expensive to train.  Fortunately, several
large pretrained diffusion models are publicly
available~\cite{StableDiffusion,DeepFloyd,VonPlaten:2022:Diffusers},
which have subsequently been adapted to perform novel tasks.  There
exists a wide variety of adaptation methods, but some of the most
commonly used ones are model finetuning and Low-Rank Adaptation
(LoRA)~\cite{Hu:2021:LRA}.  Finetuning (\ie, continuing to train the
model on a new task) requires carefully designed training sets to
avoid overfitting~\cite{Biderman:2024:LLL,Shuttleworth:2025:LFF}.
LoRAs, on the other hand, mitigate overfitting by keeping the original
weights $\W$ of linear layers, and modeling an offset $\Delta\W$ by a
low-rank approximation which furthermore regularizes the adaptation.
Formally, given the frozen linear weights $\W \in \R^{\M \times \N}$ of
a layer, the adapted weights are then expressed as: $\W + \Delta \W$,
with $\Delta \W = \A\B$, $\A \in \R^{\M \times r}$, and
$\B \in \R^{r \times \N}$, forming a low rank approximation
($r <\!< \M, \N$) of the parameter offsets.  While other regularization
techniques (\eg, weight decay and drop-out) and adaptation methods
(\eg, Elastic Weight Consolidation~\cite{Kirkpatrick:2017:OCF},
Synaptic Intelligence~\cite{Zenke:2017:CLT}, Adaptor
Layers~\cite{Houlsby:2019:PET,Lin:2020:EVG}) exist, LoRAs have the
added advantage of requiring little storage overhead or increased
latency.  However, model adaptation does not alter the number of input
and output channels.  Teamwork keeps LoRA's compactness and
efficiency, while at the same time it adds the ability to expand the
number of coordinated input and output channels.

\paragraph{Input Expansion}
A lightweight strategy to increase the number of input channels (\eg,
to add additional conditions to the diffusion model) is to expand the
input head with a zero-convolution and subsequently finetune the model
(\autoref{fig:archs}.a). Zero-convolution head expansion has been
used, for example, to add conditions to an unconditional SVBRDF
prediction model~\cite{Sartor:2023:MGD}, and to include additional
input maps in a neural shading network~\cite{Zeng:2024:RGB}.  However,
finetuning increases the risk of
overfitting~\cite{Biderman:2024:LLL,Shuttleworth:2025:LFF}.
ControlNet~\cite{Zhang:2023:ACC} offers an elegant alternative that
reduces the risk of overfitting by leaving the base model unchanged,
and instead ControlNet adds a parallel model (with identical
architecture) that provides weight offsets at each layer of the
diffusion model (\autoref{fig:archs}.b).  A variant of ControlNet is
ControlLoRA~\cite{Wu:2023:ControlLora} where the weights of the
ControlNet are adapted with LoRAs. GLIGEN~\cite{Li:2023:GOS} also
keeps the base model frozen, but adds a gated self-attention layer
between the self-attention and cross-attention in the transformer
blocks.  While suitable for input channel expansion, neither
ControlNet, nor ControlLoRA, nor GLIGEN can perform output channel
expansion.

\paragraph{Output Expansion}
Expanding the number of output channels of a pretrained diffusion
model without full retraining remains an open problem and
task-specific solutions have been introduced.  Recent work in
leveraging pretrained diffusion models for intrinsic image
decomposition~\cite{Zeng:2024:RGB,Luo:2024:IDJ} (\ie, decomposing an
image in different reflectance components such as albedo, roughness,
normals and irradiance) introduced specialized prompt keywords to
direct the diffusion model to generate the requested output.  While
each output is generated using the same model weights, there is no
explicit coordination between the inference processes of different
outputs, and coherence is implicitly encouraged by using the same seed
for each input plus a strong conditioning on the input image.  An
alternative strategy is to leverage multiple (differently) adapted
instances of a pretrained base model that each produce part of the
desired output channels. However, to ensure that all instances
generate a coherent output, some form of coordination between the
different models is needed -- without coordination each model operates
independently, cf. batching (\autoref{fig:archs}.c).
Hertz~\etal~\shortcite{Hertz:2023:SAI} share attention between the
different instances to generate multiple images with a consistent
style. Recently, Zhang~\etal~\shortcite{Zhang:2024:TIL} employed joint
attention to simultaneously generate a fore and background layer
(\autoref{fig:archs}.d).  While joint attention is a very flexible and
powerful coordination strategy, it incurs a significant (quadratic)
computational overhead cost in terms of model instances.  In contrast,
the coordination between different teammates in Teamwork does not
incur any computational overhead compared to the cost of adapting each
diffusion model instance with regular LoRA.

\paragraph{Multi-diffusion Inference}
Multidiffusion~\cite{Bar:2023:MDF} interleaves multiple diffusion
models to synthesize high resolution images and panoramas.
Multidiffusion does not explicitly coordinate between the different
diffusion instances, but instead relies on implicit coordination by
spatially overlapping each models' output and by interleaving and
combining the denoising steps.  Hence, it does not support output
expansion, only domain extension.  Furthermore, multidiffusion only
supports coordination with (spatially) nearby diffusion instances.  In
contrast, Teamwork supports channel expansion, and it provides
explicit support for coordination between all diffusion instances.

A common strategy in
video-diffusion~\cite{Ho:2022:VDM,Gupta:2025:PVG,Girdhar:2023:EVF,Blattmann:2023:ALH,Singer:2023:MAV}
is to run an image diffusion model per-frame and coordinate synthesis
between frames by augmenting the diffusion model with a combination of
3D convolutions (or factored 2D spatial and 1D temporal convolutions)
and temporal attention. However, due to the computational overhead of
temporal attention, these models often synthesize a few key frames and
then interpolate the in-between frames.  Lumiere~\cite{Bar:2024:LST}
synthesizes a whole video at once by using a cascading architecture
that down/upsamples both spatially and temporally.  A key difference
between video diffusion and Teamwork is that video diffusion performs
the same task for each frame, whereas Teamwork performs semantically
different tasks per diffusion instance.  Furthermore, due to the
memory efficiency of the low-rank approximation, Teamwork scales
better in the number of output channels.

\ignorethis{
Teamwork is also related to multi-task learning where multiple tasks
are learned in parallel. A key difference is that multi-task learning
is only concerned with running multiple tasks in parallel during
training in order to improve inference of each separate task (\ie,
tasks are not run in parallel during inference).  Therefore, we only
focus on methods that expand LoRA for multi-task learning; we refer to
the survey by Zhang and Yang~\shortcite{Zhang:2022:ASM} for a more
in-depth overview of multi-task learning.  A common strategy in
multi-task learning is to formulate it as a mixture of experts
(MoE). LoRA-based multi-task methods express each expert as a low-rank
adaptation of a common pretrained base model:
$\Delta \W_t = \sum_i^n \omega_{t,i} \A_i \B_i$, where $n$ is number
of experts, and the weights $\omega_{t,i}$ express the importance of
the $i$-th expert for the $t$-th task~\cite{Liu:2024:WMM}.  Similarly,
Ponti~\etal~\shortcite{Ponti:2023:CPE} use a MoE and only allow binary
activations. Wang~\etal~\shortcite{Wang:2023:MDL}, also optimizes the
weights, but differ in the initialization of $\B_i$. Instead of
initializing $\B_i$ to zero as common in LoRAs, Wang~\etal~ set the
initial weights $\omega_{t,i}$ to zero, and use Kaiming-uniform
initialization for $\B_i$ (as well as $\A_i$).  Finally,
Yang~\etal~\shortcite{Yang:2024:MLR} follows a slightly different
form; while each experts' $\B_i$ is still shared between tasks, each
$\A_t$ is unique per task, and the weight is an $r \times r$ learned
mixture matrix.  All these multi-task learning methods aim to share
information during training of the different tasks, while retaining
the ability to run inference on each task with the same input
separately after training, and hence do not share information during
inference.  In contrast, Teamwork shares information between the
different 'tasks' (\ie, output channels) during both training and
inference.
}

\section{Method}
\label{sec:method}

Our goal is to widen the number of output or input channels of a given
pretained diffusion model such as Stable
Diffusion~\cite{StableDiffusion,StableDiffusionXL,StableDiffusion3},
Deep Floyd~\cite{DeepFloyd}, or Flux~\cite{Flux1}, as well as adapt
the model for a given task.  Modifying the architecture of the base
diffusion model typically destroys much of the embedded knowledge.
Therefore, we retain the original architecture, and instead run
$\T = \lceil c / 3 \rceil$ adapted instances of a base diffusion
model, called \emph{teammates}, that coordinate their outputs for $c$
output/input channels.  We first explain the basis idea for widening
and adapting the output channels, and extend this idea to expanding
the input channels in~\autoref{sec:inputexpansion}.

\subsection{Output Expansion}
\label{sec:outputexpansion}
Similar as in LoRA~\cite{Hu:2021:LRA}, we focus on adapting the linear
layers of the diffusion model. Denote $\W \in \R^{\M \times \N}$ the
frozen weights of a linear layer in the base diffusion model, and
$\H = [ \H_1 \dots \H_\T ], \H_i \in \R^\M$ and
$\X = [ \X_1 \dots \X_\T ], \X_i \in \R^\N$ are the concatenated output
and input feature vectors respectively of $\T$ instances of the linear
layer. We can then compactly formulate the combined linear layer as:
\begin{equation}
  \H = %
  \begin{bmatrix}
    \W &    &     \\
       & \ddots & \\
       &    &  \W \\
  \end{bmatrix} %
  \X,
\label{eq:parallel_noadapt}
\end{equation}
which is equivalent to batching $\T$ instances of a diffusion model
(\autoref{fig:archs}.c).  Each instance is oblivious to the others,
which is expressed by the block diagonal structure of the combined
linear weight matrix in~\autoref{eq:parallel_noadapt} and thus each
$\X_i$ only affects $\H_j$ when $i = j$. Consequently, the output of
each instance is independently generated from the outputs of the other
instances.

To adapt the joint model in~\autoref{eq:parallel_noadapt} for output
channel expansion, we need (1) to adapt the behavior of each instance
(\ie, different teammates should output a different subset of the
output channels), and (2) ensure that the outputs of different
teammates are coherent (\eg, similar to how there is coherence between
R, G and B channels in an image, so is there often coherence between
different channels in multi-channel data such as SVBRDFs or intrinsic
images).  We propose to achieve both goals jointly by addition of an
offset weight matrix:
\begin{equation}
  \H = \mathtt{block}(\W) \X + \Delta\W \X,
\label{eq:teamwork}
\end{equation}
where $\Delta\W \in \R^{\T \M \times \T \N}$ are the adaptation offset
weights. The matrix $\Delta\W$ is very large, and an efficient
encoding is desired.  Inspired by LoRA~\cite{Hu:2021:LRA}, we employ a
low-rank representation:
\begin{equation}
  \Delta \W = \begin{bmatrix} \A_1 \\ \vdots \\ \A_\T \end{bmatrix} \begin{bmatrix} \B_1 & \dots & \B_\T \end{bmatrix}.
  \label{eq:deltaW}
\end{equation}
with low rank factors $\A_i \in \R^{\M \times r}$ and
$\B_i \in \R^{r \times \N}$ respectively.

To gain better insight in \autoref{eq:teamwork}, we contrast it to
applying regular LoRA to each of the $\T$ teammates.  In the latter
case, the combined weight matrix becomes:
$\Delta\W = \mathtt{block_i}{(\A_i\B_i)}$. Although uniquely adapted,
each $\H_i$ is again only affected by $\X_j$ when $i=j$.  Hence,
similar to batching (\autoref{fig:archs}.c), each teammate operates in
isolation and it is oblivious to what the other teammates are doing.
In contrast, our Teamwork framework forms a low-rank representation of
the \emph{whole} offset matrix $\Delta\W$. Crucially, because the
resulting $\Delta\W$ is not block-diagonal anymore, each output $\H_i$
is now affected by all inputs $\X_j$, allowing for feature information
to flow between different teammates (\ie, coordination).

We can also draw parallels between Teamwork and Joint
Attention~\cite{Zhang:2024:TIL}.  Joint attention enables coordination
between different instances of a(n adapted) diffusion model by
replacing the self-attention layers with joint-attention layers across
all models (\autoref{fig:archs}.d).  However, Joint Attention suffers
from a quadratic computation complexity with respect to the number of
models (as well as the resolution). Furthermore, Joint Attention only
offers coordination and a separate adaptation method is needed (\eg,
regular LoRA on each instance).  Teamwork is in some sense the dual of
Joint Attention; instead of exchanging information via the attention
layers, Teamwork instead shares information via the linear features of
the model (\autoref{fig:archs}.e).  This offers three advantages: (1)
coordination and adaption are performed jointly, (2) the number of
coordination layers quadruples (\ie, four linear layers per attention
layer), and (3) the computational cost is identical to applying a
regular LoRA to each instance (\ie, linear with the number of
teammates; see~\autoref{sec:practical}).

\subsection{Input Expansion}
\label{sec:inputexpansion}
We can also leverage Teamwork to widen the number input channels (\ie,
add condition images).  Similar as with output expansion, we add a
teammate for each triplet of input conditions; we differentiate
between both types of teammates as \emph{input-teammates} and
\emph{output-teammates}.  The key idea is that diffusion models
extract sophisticated features that are not only useful for image
synthesis, but that are also meaningful for conditioning the diffusion
process; these features are present irrespective of the degree of
noise in the input.  Thus input-teammates do not function as diffusion
models, but as vision models. Therefore, we pass the noise-free input
image at \emph{every} diffusion step instead of starting from noisy
latents, and propagate the resulting features.  This is equivalent to
ignoring the output of the input-teammates at every diffusion step.
Consequently, no loss is defined on the input-teammates.  However,
this does not mean that the input-teammates are not adapted; the loss
from the output-teammates is still propagated via the dense offset
weights $\Delta\W$ to the
input-teammates.

\subsection{Dynamic Activation of Teammates}
\label{sec:enable}
Teamwork also supports dynamic (de)activation of teammates, both
during inference and training.  The latter is particularly useful when
training on heterogenous datasets that have different subsets of the
channels available. Similar to RGB$\rightarrow$X~\cite{Zeng:2024:RGB},
rather than selecting the largest common subset for training, we can
during training dynamically activate the channels present for each
training exemplar, thereby maximally leveraging the information in the
combined training set.  Practically, dynamic activation is achieved by
only including the low rank factors $\A_i$ and $\B_i$ for active
teammates (and corresponding $\X_i$) when evaluating $\Delta\W\X$
(\autoref{eq:deltaW}).

\subsection{Practical Considerations}
\label{sec:practical}
\paragraph{Materialization}
Materializing $\W+\Delta\W$ in regular LoRA is common practice because
it slightly reduces computational evaluation costs from
$\mathcal{O}(\M \N + r (\M + \N))$ to $\mathcal{O}(\M \N)$.  However, for
Teamwork with $\T$ teammates, we can exploit that the frozen component
is a block-matrix, yielding an unmaterialized evaluation cost of
$\mathcal{O}(\T \M \N + \T r (\M + \N))$, compared to
$\mathcal{O}(\T^2 \M \N)$ for evaluation on a materialized
$\W+\Delta\W$, and hence materialization is more expensive for modest
Teamwork-rank.  Furthermore, the memory cost of materialization for
Teamwork is more significant due to the dense off-diagonal blocks.

\paragraph{Batch Trick}
To simplify implementation of Teamwork, we exploit the inherent
capability of batching to run multiple instances of a diffusion model
simultaneously, and expand it by performing Teamwork-aware (low-rank)
operations across the batch dimensions. As a result, linear layers
perform Teamwork-aware operations across batch-dimension, while other
layers broadcast across teammates without performing any coordination.
However, implementing this batch trick limits the micro-batch size to
$1$, necessitating the use of gradient accumulation during training.
Practically, we found that gradient accumulation did not pose a major
disadvantage as common team sizes ($5$-$10$) can easily saturate a
single GPU's memory, necessitating distributed gradient accumulation.

\section{Results}
\label{sec:results}
Teamwork is designed as a general framework for input and output
expansion of pretrained diffusion models.  To demonstrate the
versatility and flexibility of our framework, we apply Teamwork to
five different generative and inverse rendering tasks (inpainting,
single image SVBRDF estimation, intrinsic image decomposition, neural
shading, and intrinsic image synthesis) that require an
expansion of the input and/or output channels.  All training (and
timing) is performed on a single NVIDIA A40 with $48$GB of VRAM. We
use the Prodigy optimizer~\cite{Mishchenko:2023:PEA} to train Teamwork
using $16\times$ gradient accumulation and using a cosine learning
rate schedule with the standard loss for which the base model was
trained (\ie, diffusion loss for Stable Diffusion
XL~\cite{StableDiffusionXL} or flow-matching loss for Stable Diffusion
3~\cite{StableDiffusion3} and Flux~\cite{Flux1}).

\begin{figure}
  \begin{center}%
    \includegraphics{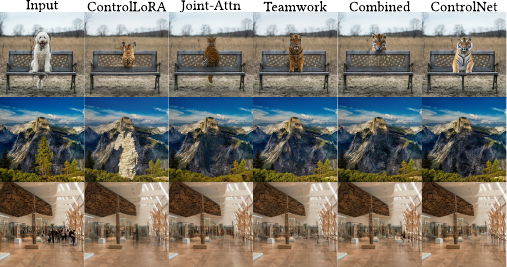}%
  \end{center}%
  \caption{Teamwork performs qualitative similarly to different Stable
    Diffusion 3 based inpainting methods.}
  \label{fig:inpainting}
\end{figure}

\paragraph{Inpainting}
We employ Stable Diffusion 3~\cite{StableDiffusion3} as the base
diffusion model and adapt it for the generative task of
inpainting~\cite{Rombach:2022:HRI}.  We use three teammates; two
input-teammates: one for the image and one for the mask (copied to the
three RGB channels), and one output-teammate for the resulting
inpainted image.  We compare Teamwork to different input expansion
techniques such as ControlLoRA~\cite{Wu:2023:ControlLora}, Joint
Attention~\cite{Zhang:2024:TIL} for coordinating between separately
LoRA-adapted teammates, and a variant that employs both Joint
Attention and Teamwork for coordination as both methods can be
employed simultaneously. All models are trained on $64$k images from
the PixelProse dataset~\cite{Singla:2024:FPP} at $1024$ pixel
resolution. Training took $31$h for ControlLoRA and Teamwork, and
$100$h for the Joint Attention and combined model due the
$\mathcal{O}(T^2)$ time-complexity (versus $\mathcal{O}(T)$ for
Teamwork).  In addition, we also compare against a pretrained
ControlNet-based inpainting variant of Stable Diffusion
3~\cite{StableDiffusionInpainting} which was trained with a larger
batch size ($192$) and on a much larger training set of
$3,\!840,\!000$ images from Laoin2B~\cite{Laion2b} and a proprietary
dataset, and thus likely required significantly more resources for
training.  We quantitatively compare (\autoref{tab:inpainting}) the
different models on a random test split of the PixelProse dataset for
four different metrics: SI-FID~\cite{Shaham:2019:SLG} ($2$nd column;
lower is better), CLIPScore~\cite{Hessel:2021:CAR} on the inpainted
image of the reference ($3$rd column; higher is better) and with
respect to the input prompt ($4$th column; higher is better), and CLIP
IQA~\cite{Wang:2023:ECA} on the inpainted model (last column; higher
is better).  All methods perform similarly, both quantitatively
(\autoref{tab:inpainting}) and qualitatively
(\autoref{fig:inpainting}), demonstrating that Teamwork offers a
viable and competitive alternative for input channel expansion with
reduced training costs.

\begin{table}
  \centering
  \caption{Teamwork performs quantitatively similarly compared to
    other Stable Diffusion 3 based inpainting methods on a random test
    split from the PixelProse dataset~\cite{Singla:2024:FPP} for four
    different metrics: SI-FID~\cite{Shaham:2019:SLG},
    CLIPScore~\cite{Hessel:2021:CAR} with respect to the text prompt
    and input image, and IQA~\cite{Wang:2023:ECA}.}
  \label{tab:inpainting}
  \footnotesize
  \begin{tabular}{r|cccc}
    Method & SI-FID $\downarrow$ & CLIPScore (Img) $\uparrow$ & CLIPScore (Txt) $\uparrow$ & CLIP IQA $\uparrow$ \\
    \hline
    ControlNet      & \UL{13.987} & \BF{90.18} & \UL{31.53} & \BF{0.530} \\
    ControlLoRA     & 17.856      & 86.97      & 31.27      & 0.469      \\
    Joint-Attn.     & 14.147      & \UL{89.94} & 31.02      & 0.486      \\
    Teamwork        & \BF{13.236} & 89.65      & 31.05      & 0.490      \\
    Combined        & 16.473      & 89.88      & \BF{31.59} & \UL{0.515} \\
  \end{tabular}
\end{table}

\paragraph{Single Image SVBRDF Estimation}
SVBRDF estimation aims to recover spatially-varying reflectance
distribution (SVBRDF) parameters (\eg, diffuse albedo, specular
albedo, specular roughness, and surface normals) from a single
photograph of a planar material sample under controlled or
uncontrolled lighting.  SVBRDF estimation is traditionally formulated
as an inverse rendering problem. Single image SVBRDF estimation is an
inherently underconstrained problem, and generative diffusion-based
SVBRDF estimation methods (MatFusion~\cite{Sartor:2023:MGD} and
ControlMat~\cite{Vecchio:2023:CAC}) have recently been shown to
address the resulting ambiguities better than purely regressive
techniques.  However, to support the expanded number of output
channels ($10$ channels in total) these prior diffusion-based SVBRDF
solutions are trained from scratch.

\begin{figure}
  \begin{center}%
    \includegraphics{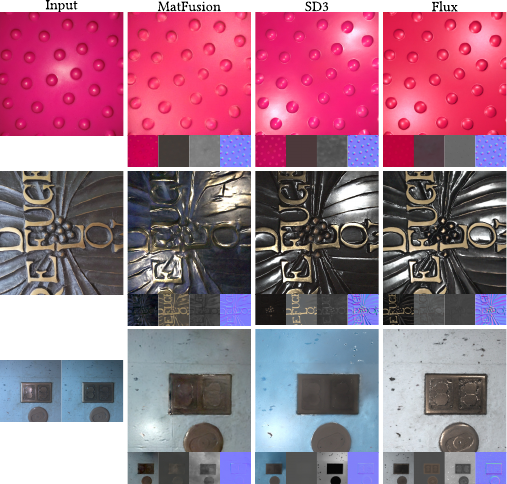}%
  \end{center}%
  \caption{Qualitative comparison of SVBRDFs estimated from real-world
    captured photographs of Teamwork variants and
    MatFusion~\cite{Sartor:2023:MGD} under colocated ($1$st row),
    environment ($2$nd row), and flash/no-flash ($3$rd row) lighting.}
  \label{fig:realsvbrdf}
\end{figure}

We train three Teamwork variants for SVBRDF estimation under three
different lighting conditions mirroring
MatFusion~\cite{Sartor:2023:MGD}: colocated flash lighting, unknown
environment lighting, and flash/no-flash image pairs.  Each Teamwork
model includes at minimum $3$ input channels (photograph under target
lighting) and $12$ output channels ($3$ for diffuse albedo (gamma
$2.2$ encoded), $3$ for specular albedo (gamma $2.2$ encoded), $3$ for
normals, and $3$ for roughness encoded as a gray-scale image), \ie,
$4$ output teammates.  For the colocated model, we follow MatFusion
and add a second input-teammate containing the halfway vector between
the light and view vector per pixel, yielding a total of $2$ input and
$4$ output-teammates.  For the environment lighting variant, we add an
extra output-teammate for the shading-only image that corresponds to
(the normalized) ratio of the photograph over the sum of the diffuse
and specular albedo, yielding a total of $1$ input and $5$ output
teammates.  Finally, the flash/no-flash model features $2$
input-teammates (one for the flash and one for the no-flash
photograph), and the same $5$ output-teammates as the environment
lighting variant. We train each Teamwork variant on $64$k exemplars
from the MatFusion SVBRDF training dataset.  While MatFusion is
limited to $256 \times 256$ resolution due to the computational cost
of training a diffusion model from scratch, we train our models at
$512 \times 512$ resolution.

\begin{figure*}
  \begin{center}%
    \includegraphics{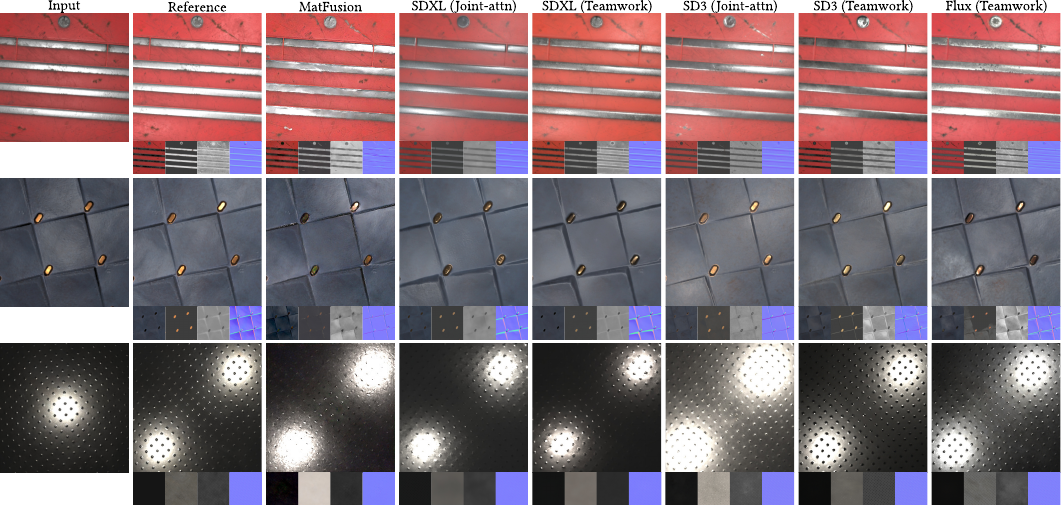}%
  \end{center}%
  \caption{Qualitative comparison of Teamwork variants and
    MatFusion~\cite{Sartor:2023:MGD} on simulated captures under
    colocated lighting for three synthetic SVBRDFs. For each SVBRDF we
    show a rerendering under novel lighting and the estimated diffuse
    albedo, specular albedo, roughness, and normal maps.}
  \label{fig:svbrdf}
\end{figure*}

\begin{table}
  \centering
  \caption{Quantitative comparison of different Teamwork variants for
    single image SVBRDF estimation under colocated, environment, and
    flash/no-flash lighting.  For reference, we compare against the
    diffusion-based MatFusion~\cite{Sartor:2023:MGD} models using the
    average RMSE on the estimated parameter maps and the average
    LPIPS~\cite{Zhang:2018:TUE} rerender error under $128$ randomly
    placed point lights on the MatFusion test set.}
  \label{tab:svbrdf}
  \footnotesize
  \begin{tabular}{c|ll|ccccc}
    Lighting & Base & Coord. & Render & \multicolumn{4}{c}{RMSE} \\
             & Model& Method & Error  & Diff. & Spec. & Rough. & Norm. \\
    \hline
    \multirow{6}{*}{Coloc.}   & MatFus.       & N.A.   & 0.214 & 0.071 & 0.0958 & 0.115 & 0.0562 \\
                              & SDXL          & Attn   & 0.216 & \BF{0.0471} & \UL{0.0861} & \BF{0.105} & \UL{0.0454} \\
                              & SD3           & Attn   & 0.214 & 0.0682 & 0.117 & 0.145 & 0.0481 \\
                              & SDXL          & Team.  & 0.204 & \UL{0.0572} & 0.0876 & \BF{0.105} & \BF{0.0416} \\
                              & SD3           & Team.  & \BF{0.188} & 0.0609 & \BF{0.0835} & 0.12 & 0.0464 \\
                              & Flux          & Team.  & \UL{0.195} & 0.0653 & 0.101 & 0.141 & \UL{0.0454} \\
    \hline
    \multirow{3}{*}{Environ.} & MatFus.       & N.A.   & 0.337 & 0.197 & 0.133 & 0.258 & 0.087 \\
                              & SD3           & Team.  & \BF{0.266} & \BF{0.143} & \BF{0.105} & \BF{0.183} & \BF{0.0538} \\
                              & Flux          & Team.  & \UL{0.273} & \UL{0.17} & \BF{0.105} & \UL{0.2} & \UL{0.0551} \\
    \hline
    Flash /                   & MatFus.       & N.A.   & 0.3709      & 0.187      & 0.120 & \UL{0.143} & \UL{0.062} \\
    No-flash                  & SD3           & Team.  & \bf{0.2518} & \bf{0.150} & \bf{0.100} & \bf{0.116} & \bf{0.050} \\
    pair                      & Flux          & Team.  & \UL{0.3184} & \UL{0.159}      & \UL{0.112} & 0.235 & 0.065
  \end{tabular}
\end{table}

We demonstrate Teamwork's flexibility by using different base
diffusion models (\ie, Stable Diffusion XL~\cite{StableDiffusionXL},
Stable Diffusion 3~\cite{StableDiffusion3}, and
Flux~\cite{Flux1}). Qualitative comparison of the three variants
against their respective MatFusion~\cite{Sartor:2023:MGD} counterparts
are shown in~\autoref{fig:realsvbrdf},~\autoref{fig:svbrdf}, and the
supplemental material. \autoref{tab:svbrdf} quantitatively compares
the different methods against MatFusion, as well as selected variants
that use Joint Attention for coordination instead.  From the
quantitative comparison, we can see that the Teamwork variants
outperform MatFusion as well as the Joint Attention variants in terms
of rerender error, and that all models perform competitively on
accuracy for each of the reflectance components.  Not only do the
Teamwork variants perform better than MatFusion models despite being
trained on significantly fewer training exemplars and using fewer
training iterations (only $4,\!000$ versus $672,\!000$ for MatFusion),
they are also significantly faster to train: $21$/$18$/$85$ GPU hours
for the Stable Diffusion XL, Stable Diffusion 3, and Flux base-model
respectively at $512$ resolution versus $\sim\! 400$ GPU hours for
MatFusion (refinement time from the MatFusion base model) at $256$
resolution; a $5\!\sim\!20\times$ speed-up.  For reference, the Joint
Attention variant required $22$/$41$ GPU hours to train for the Stable
Diffusion XL and Stable Diffusion 3 base models respectively. Note
that the impact on the training cost for Stable Diffusion XL with
Joint Attention versus Teamwork is less significant because Stable
Diffusion XL features relatively few attention layers compared to
convolutional layers. More modern diffusion architectures, such as
Stable Diffusion 3 and Flux, depend almost exclusively on attention
layers, and hence incur a more significant training cost increase for
Joint Attention.

\begin{table*}
  \def\NMC{\multicolumn{2}{c}{\xmark}}
  \def\N{\xmark}
  \centering
  \caption{Quantitative comparison of Intrinsic Image
    Diffusion~\cite{Kocsis:2023:IID},
    RGB$\rightarrow$X~\cite{Zeng:2024:RGB} (pretrained and
    InteriorVerse retrained variants) and a Stable Diffusion 3 based
    Teamwork model (a variant trained on a heterogeneous training set
    and a variant trained only on InteriorVerse) over the
    InteriorVerse test set.  To compensate for the albedo-intensity
    ambiguity, we first apply a least-squares optimization to find an
    optimal scale per channel before computing the respective errors
    for each method.}
  \label{tab:indoor}
  \footnotesize
  \begin{tabular}{l|r|ccccccccccc}
    & \multirow{2}{*}{Method}  & \multicolumn{2}{c}{Diffuse} & \multicolumn{2}{c}{Specular} & \multicolumn{2}{c}{Roughness} & \multicolumn{2}{c}{Albedo} & \multicolumn{2}{c}{Shading} & Normal \\
                         &                                         & RMSE  & LPIPS & RMSE  & LPIPS & RMSE  & LPIPS & RMSE  & LPIPS & RMSE  & LPIPS & RMSE  \\
    \hline
    \multirow{3}{*}{I}   & Kocsis~\etal                             & 0.135 & 0.205 & \UL{0.126} & 0.242 & 0.232 & 0.384 & \UL{0.110} & \UL{0.166} & 0.144 & 0.230 & \N \\
                         & RGB$\rightarrow$X                        & 0.229 & 0.518 & 0.272 & 0.576 & 0.304 & 0.654 & 0.148 & 0.378 & 0.065 & 0.137 & 0.196 \\
                         & Teamwork                                 & 0.176 & 0.248 & 0.151 & 0.288 & 0.291 & 0.424 & 0.140 & 0.208 & 0.064 & 0.140 & 0.121 \\
    \hline
    \multirow{3}{*}{II}  & RGB$\rightarrow$X (InteriorVerse)        & 0.165 & 0.300 & 0.143 & 0.293 & 0.249 & 0.371 & 0.139 & 0.273 & 0.073 & 0.211 & 0.133 \\
                         & RGB$\rightarrow$X SD3 (InteriorVerse)    & 0.141 & 0.223 & 0.132 & 0.274 & \UL{0.212} & \UL{0.314} & 0.127 & 0.203 & 0.058 & 0.159 & 0.116 \\
                         & Teamwork (InteriorVerse)                 & \BF{0.116} & \BF{0.156} & \BF{0.119} & \BF{0.197} & \BF{0.189} & \BF{0.259} & \BF{0.107} & \BF{0.136} & \BF{0.045} & \BF{0.097} & \BF{0.093}\\
    \hline
    \multirow{5}{*}{III} & No Coordination                          & 0.205 & 0.353 & 0.169 & 0.370 & 0.329 & 0.508 & 0.170 & 0.334 & 0.083 & 0.225 & 0.171 \\
                         & Sequential Teamwork                      & 0.236 & 0.372 & 0.145 & 0.359 & 0.313 & 0.612 & 0.190 & 0.265 & 0.070 & 0.172 & 0.141 \\
                         & InteriorVerse-restricted Teamwork        & \UL{0.128} & \UL{0.184} & 0.150 & \UL{0.229} & 0.243 & 0.321 & 0.125 & 0.167 & \UL{0.054} & \UL{0.119} & \UL{0.107} \\
                         & Teamwork (Dropout)                       & 0.195 & 0.310 & 0.154 & 0.293 & 0.321 & 0.461 & 0.162 & 0.261 & 0.091 & 0.189 & 0.135 \\
                         & Sequential Teamwork (Dropout)            & 0.221 & 0.293 & 0.152 & 0.292 & 0.303 & 0.450 & 0.173 & 0.248 & 0.066 & 0.155 & 0.132 \\
  \end{tabular}
\end{table*}

\begin{figure*}
  \begin{center}%
    \includegraphics{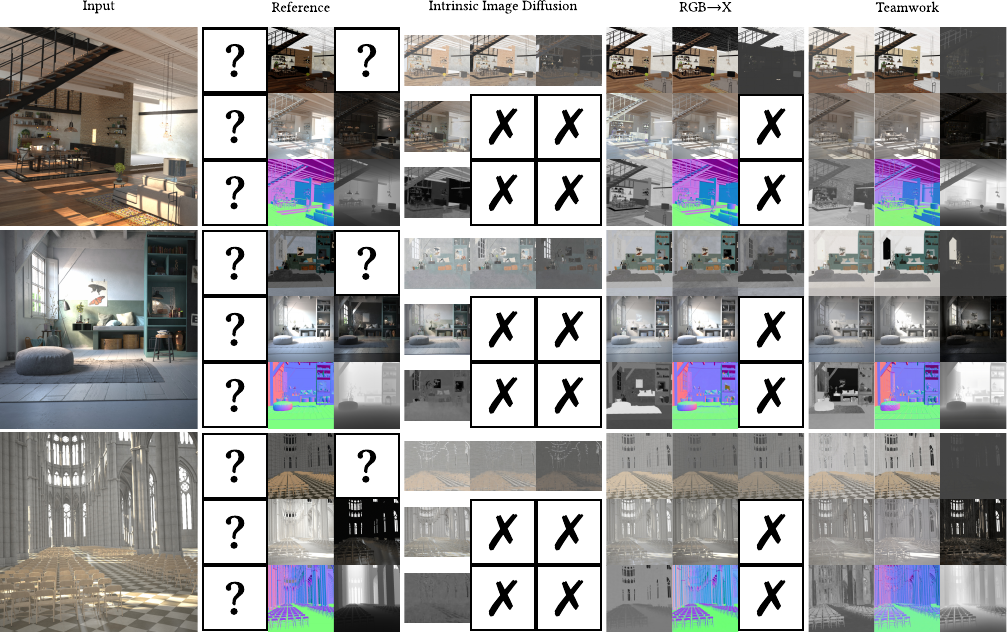}%
  \end{center}%
  \caption{Qualitative comparison (on examples from
    HyperSim~\cite{Roberts:2021:HPS}) of the pretrained Instrinsic
    Image Diffusion~\cite{Kocsis:2023:IID} and pretrained
    RGB$\rightarrow$X~\cite{Zeng:2024:RGB} against a Stable Diffusion
    3 based Teamwork variant trained on the heterogeneous training
    set. For each method the resulting intrinsic components (if
    available) are organized as: ($1$st row): summed albedo, diffuse
    albedo, and specular albedo; ($2$nd row): shading, diffuse
    shading, specular residual; and ($3$rd row): roughness, normals,
    and depth.}
  \label{fig:intrinsic}
\end{figure*}

\paragraph{Intrinsic Image Decomposition}
Intrinsic image decomposition aims to separate an image into various
shading components, hence it constitutes a one-to-many image
translation.  While originally designed to decompose into just two
components (\ie, reflectance and shading), more recently methods
consider an expanded definition of intrinsic
components~\cite{Kocsis:2023:IID,Zeng:2024:RGB}; we follow this later
inverse rendering view of intrinsic decomposition.  We adapt Stable
Diffusion 3~\cite{StableDiffusion3} to include $1$ input-teammate (for
the photograph to be decomposed) and $9$ different output-teammates at
$1024$ resolution for: diffuse albedo, specular albedo, the summed
(diffuse and specular) albedo, specular roughness, normals, depth,
diffuse shading, (diffuse plus specular) shading, and the specular
residual defined as the difference between the image and the diffuse
albedo times the diffuse shading.

To demonstrate Teamwork's ability to support heterogeneous training
datasets, we randomly select $256$k training exemplars from:
InteriorVerse~\cite{Zhu:2022:LBI}, HyperSim~\cite{Roberts:2021:HPS},
CGIntrinsics~\cite{Li:2018:CBI} (using the CGIntrinsic's albedo as
total reflectance), Infinigen~\cite{Raistrick:2023:IPW}, and
renderings of random objects selected from the ABC
dataset~\cite{Koch:2019:ABC} textured with random MatFusion
SVBRDFs~\cite{Sartor:2023:MGD} and lit by a random light probe from
\url{https://polyhaven.com/hdri}; not all reflectance components are
present in each dataset (see supplemental material for a summary of
the available components).  The resulting Teamwork model
performs comparably (\autoref{tab:indoor}, Part I) to the
publicly-available Intrinsic Image Diffusion~\cite{Kocsis:2023:IID}
and RGB$\rightarrow$X~\cite{Zeng:2024:RGB} models over the
InteriorVerse test set.  \autoref{fig:intrinsic} further
qualitatively demonstrates intrinsic image decompositions on three
selected images from HyperSim~\cite{Roberts:2021:HPS}; see
supplemental material for additional decompositions on other datasets.

While informative, the previous comparison is not conclusive, as all
three models are trained on different datasets.  Therefore, we retrain
RGB$\rightarrow$X and Teamwork on $256$k random samples drawn from the
InteriorVerse training set only (corresponding to $\sim\!\!6$ epochs
over the full InteriorVerse training set); the pretrained Intrinsic
Image Diffusion model~\cite{Kocsis:2023:IID} is already exclusively
trained on InteriorVerse and thus not retrained. Furthermore, as
RGB$\rightarrow$X uses Stable Diffusion 2.1 as a base-model while
Teamwork leverages the more powerful Stable Diffusion 3 as a
base-model, we also train an Stable Diffusion 3 variant of
RGB$\rightarrow$X.  As can be seen in~\autoref{tab:indoor} (Part II)
the retrained RGB$\rightarrow$X models both show improvements compared
to its more general pretrained counterpart over the InteriorVerse test
set with the Stable Diffusion 3 variant slightly outperforming the
Stable Diffusion 2.1 variant.  However, the performance of the
InteriorVerse-specialized Teamwork outperforms prior methods by a
significant margin.

\begin{figure}
  \begin{center}%
    \includegraphics{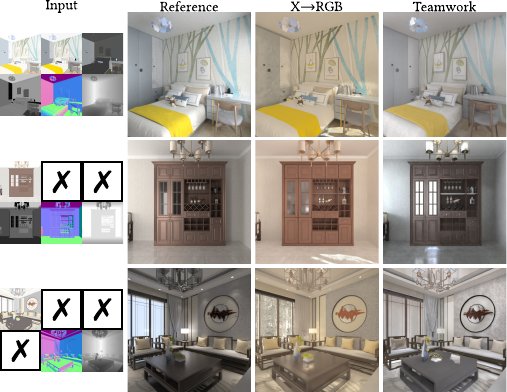}%
  \end{center}%
  \caption{Neural shading from subsets of intrinsic parameters ($1$st
    column - first row: summed albedo, diffuse albedo, specular
    albedo; second row: roughness, normals, depth). While an exact
    match to the reference ($2$nd column) is unlikely due to the
    uncontrolled lighting, Teamwork ($4$th column) produces
    qualitatively similar results to X$\rightarrow$
    RGB~\cite{Zeng:2024:RGB} ($3$rd column).}
\label{fig:shading}
\end{figure}

\paragraph{Neural Shading}
Inspired by X$\rightarrow$RGB~\cite{Zeng:2024:RGB} we also create an
intrinsic neural shader that generates from a set of intrinsic
components, a synthetic image under unknown random lighting that
conforms to the provided components.  We adapt Stable Diffusion
3~\cite{StableDiffusion3} for this task, using the same channels as
for the intrinsic image decomposition Teamwork model, except with
different input/output roles, \ie, using shading and render channels as
output ($4$ teammates) and all other intrinsic channels as input ($6$
teammates). The neural shading Teamwork model is trained with $64$k
random exemplars from InteriorVerse, HyperSim, and CGIntrinsics.  In
contrast to X$\rightarrow$RGB, we do not overload a black image to
indicate the absence of an input channel, but instead rely on
Teamwork's dynamic (de)activation of teammates.  \autoref{fig:shading}
qualitatively compares Teamwork with X$\rightarrow$RGB; a quantitative
comparison is difficult due to the unknown lighting in the synthesized
images.  Qualitatively, we see that Teamwork performs similarly to
X$\rightarrow$RGB.

\begin{figure}[h!]
  \begin{center}%
    \includegraphics{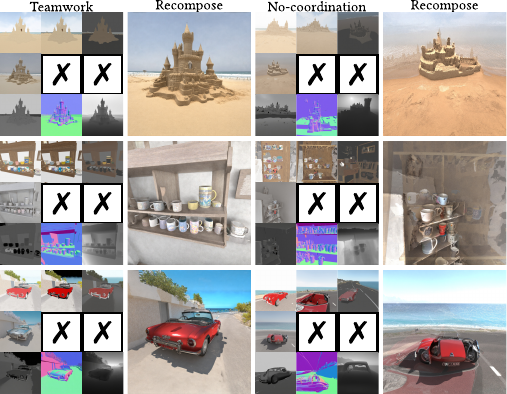}%
  \end{center}%
  \caption{Left: results from a prompt-only conditioned intrinsic
    image synthesis Teamwork model trained on $128$k exemplars from
    InteriorVerse with Gemma-3 generated prompts.  Right: results from
    models trained without coordination (and evaluated with the same
    seed) lack coherence.}
  \label{fig:intrinsicSynthesis}
\end{figure}

\paragraph{Intrinsic Image Synthesis}
As a final application, we train a novel generative Teamwork model
that directly synthesizes intrinsic components from a given prompt.
We employ the same teammates as for intrinsic image decomposition and
neural rendering, except that all $10$ are now output-teammates.  We
train this generative model on $128$k exemplars drawn from
InteriorVerse with Gemma-3~\cite{Gemma3} generated prompts.  The
intrinsic synthesis results shown in \autoref{fig:intrinsicSynthesis}
(first two columns) are qualitatively similar to the intrinsic
components from the intrinsic decomposition
in~\autoref{fig:intrinsic}.

\begin{table}
    \centering
    \caption{Quantitative ablation of Teamwork hyper-parameters on the
      Stable Diffusion 3 based environment SVBRDF estimation model
      trained with 16 accumulations per optimizer step, and using
      rank=16 and steps=4k if not otherwise specified.}
    \label{tab:ablation}
    \footnotesize
    \begin{tabular}{l|ccccc}
      Parameter & Render & \multicolumn{4}{c}{RMSE} \\
                & Error  & Diff. & Spec. & Rough. & Norm. \\
      \hline
      rank=8    & 0.286 & \UL{0.149} & 0.108 & \BF{0.175} & 0.0574 \\
      rank=16   & \BF{0.266} & \BF{0.143} & 0.105 & \UL{0.183} & \BF{0.0538} \\
      rank=64   & \UL{0.273} & 0.166 & \UL{0.101} & 0.229 & 0.0612 \\
      rank=128  & 0.279 & 0.151 & \BF{0.0991} & 0.2 & \UL{0.0557} \\
      \hline
      steps=2k & 0.286 & 0.147 & \UL{0.107} & \BF{0.179} & 0.0566 \\
      steps=4k & \UL{0.266} & \UL{0.143} & \BF{0.105} & 0.183 & \UL{0.0538} \\
      steps=8k & \BF{0.253} & \BF{0.124} & 0.108 & \UL{0.182} & \BF{0.0525} \\
    \end{tabular}
  \end{table}

\section{Discussion}
\label{sec:discusion}

\paragraph{Hyper-parameter Ablation}
To validate Teamwork's hyper-para\-me\-ter sensitivity with respect to
rank and number of training steps, we perform two $1$D parameter-scans
on the Stable Diffusion 3 based environment SVBRDF estimation model
(\autoref{tab:ablation}). We train each model with $16\times$ gradient
accumulation (in lieu of batching) with $4$k optimization steps (\ie,
$16 \times 4$k = $64$k training exemplars) and rank $16$ unless
specified otherwise.  While the optimal parameters are task specific,
we observe that Teamwork is robust across a wide range of
hyper-parameters. In general, a too low rank restricts the adaptation
capabilities whereas a too high rank offers too much freedom, and thus
increases the risk of finding a sub-optimal local minimum.  More
training steps/exemplars is generally more beneficial, albeit with
diminishing returns.

\paragraph{Teamwork vs. Joint Attention}
The Joint Attention inpainting (\autoref{tab:inpainting}) and SVBRDF
estimation (\autoref{tab:svbrdf}) variants both underperform
compared to Teamwork.  We posit that this is mainly due to two
reasons.  First, from~\autoref{fig:archs} we can see that Teamwork
contains $4\times$ as many coordination layers than Joint Attention,
and thus Teamwork can more effectively coordinate. Second, Teamwork
strongly prefers pixel-aligned input and output images. In contrast,
Joint Attention has no such built-in prior and thus needs to learn the
positional embeddings and ignore spurious correlations. We expect
Joint Attention to perform better on non-pixel aligned inputs.

\begin{figure}%
    \begin{center}%
    \includegraphics{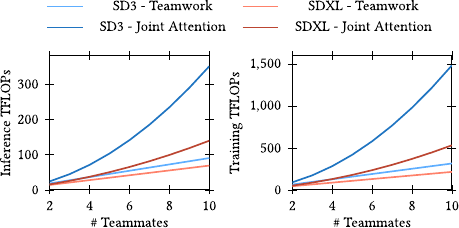}%
    \end{center}%
    \caption{Comparison of measured computational inference and
      training FLOPs (for a single diffusion step) of Teamwork
      (linear) versus Joint Attention (quadratic) for two different
      base diffusion models with respect to an increasing number of
      teammates.}
    \label{fig:flops}
\end{figure}

An additional key benefit of Teamwork over Joint Attention is
scalability with respect to resolution and number of teammates.
\autoref{fig:flops} compares inference and training FLOPs
(measured using \textsc{torch.utils.flop\_counter}) for Stable Diffusion XL and
Stable Diffusion 3 Teamwork and Joint Attention variants for an
increasing number of teammates over a single diffusion step with a
$1024$ resolution, a rank of $16$, and \textsc{bf16} datatype. The
empirically measured FLOPs confirm the theoretical linear and
quadratic complexity of Teamwork and Joint Attention respectively.

\paragraph{Importance of Coordination}
A key component of Teamwork is the coordination between teammates.  To
demonstrate the benefit and necessity of coordination, we train a
\emph{set} of Stable Diffusion 3 based intrinsic image decomposition
models using ControlLoRA for input expansion (trained on the
heterogeneous training set) that each produce one of the intrinsic
components without coordination, and evaluate the models on the
InteriorVerse test set (\autoref{tab:indoor}, Part III, first
row). Compared to the Teamwork model (\autoref{tab:indoor}, Part I),
we observe a decrease in performance, demonstrating the importance of
coordination.  However, a low error does not guarantee that the errors
between the different outputs are coherent.  To demonstrate that
Teamwork also helps to improve coherency between output channels, we
measure the joint \emph{albedo $\times$ shading} reconstruction error
with respect to the input image, yielding a $0.1127$ LPIPS and
$0.1272$ RMSE for Teamwork versus $0.2443$ and $0.2617$ respectively
for the no-coordination model.

As a final experiment to demonstrate both quality and consistency, we
train a \emph{set} of prompt-conditioned intrinsic image synthesis
networks that each produce a single component (without coordination)
and use a common seed (as in RGB$\rightarrow$X) when inferencing each
intrinsic component. \autoref{fig:intrinsicSynthesis} (last two
columns) shows that while each synthesized output is of high quality,
the maps are not mutually coherent, resulting in severe ghosting
artifacts when recomposing the intrinsic components.  In contrast, the
Teamwork generative model (first two columns) produces mutually
coherent intrinsic components.

\paragraph{Dynamic (de)activation}
To understand the impact of (de)acti\-vat\-ing teammates on the
accuracy of the model, we compare the performance of the intrinsic
image decomposition Teamwork model (trained on the heterogeneous
training set) on different subsets of (activated) teammates.  We
consider the following subsets: (a) evaluating each teammate
separately (\ie, one active teammate at the time --
\autoref{tab:indoor}, Part III, second line), (b) evaluating only
teammates restricted to components present in InteriorVerse (\ie, $7$
active output teammates -- \autoref{tab:indoor}, Part III, third
line), and (c) evaluating all $9$ output teammates jointly
(\autoref{tab:indoor}, Part I, third line).  We observe that
evaluating each component in isolation incurs a significant loss of
accuracy as this effectively disables coordination between the
teammates. However, activating only the InteriorVerse components
yields lower error than activating all components.  We posit that
Teamwork attempts to balance errors on the subsets seen during
training -- InteriorVerse is part of the training set, and thus the
Teamwork model is partially optimized for the subset of InteriorVerse
components.  To validate this thesis, we also train a Teamwork variant
with drop-out (\ie, each teammate is deactivated with a $20\%$
probability), and compare its performance on activating each component
sequentially versus all components at once (\autoref{tab:indoor}, Part
III, last two lines).  Due to drop-out, the resulting model is forced
to rely less on coordination between the teammates. Consequently the
performance gap between sequential and full inference is significantly
reduced. However, the overall performance of the drop-out trained
Teamwork is also reduced and is more similar to the no-coordination
model.

\begin{table}
  \def\N{\xmark}
  \centering
  \caption{Quantitative evaluation of the generalization capabilities
    over HyperSim~\cite{Roberts:2021:HPS} and
    Infinigen~\cite{Raistrick:2023:IPW} of Intrinsic Image
    Diffusion~\cite{Kocsis:2023:IID},
    RGB$\rightarrow$X~\cite{Zeng:2024:RGB} and a Teamwork model; the
    latter two are trained on $256$k exemplars from InteriorVerse to
    maximize the difference between training and test sets. Only
    outputs common between the different models and test sets are
    included below.}
  \label{tab:generalization}
  \footnotesize
  \begin{tabular}{r|ccc||ccc}
                       & \multicolumn{3}{c}{HyperSim}              & \multicolumn{3}{c}{Infinigen (Outdoor)}\\
    Method             & \multicolumn{2}{c}{Diffuse} & Normal      & \multicolumn{2}{c}{Diffuse} & Normal \\
                       & RMSE       & LPIPS          & RMSE        & RMSE       & LPIPS      & RMSE       \\
    \hline
    Kocsis~\etal       & \UL{0.237} & \UL{0.418}     & \N          & \UL{0.230} & 0.615      & \N         \\
    RGB$\rightarrow$X  & 0.340      & 0.484          & \UL{0.157} & \BF{0.223} & \UL{0.596} & \UL{0.276} \\
    Teamwork           & \BF{0.215} & \BF{0.356}     & \BF{0.125}  & 0.259      & \BF{0.549} & \BF{0.265} \\
 \end{tabular}
\end{table}

\begin{figure}[h!]
  \begin{center}%
    \includegraphics{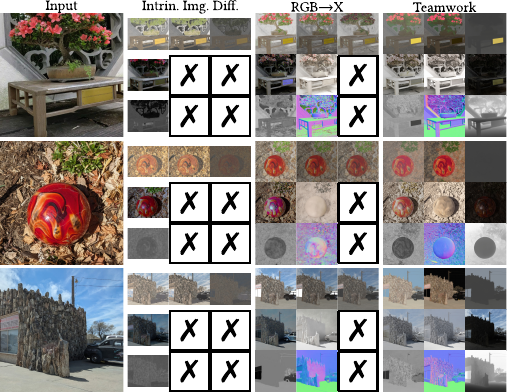}%
  \end{center}%
  \caption{Qualitative comparison of intrinsic image decompositions of
    the pretrained Intrinsic Image Diffusion~\cite{Kocsis:2023:IID},
    RGB$\rightarrow$X~\cite{Zeng:2024:RGB}, and Teamwork (trained on
    the heterogeneous data) on real-world photographs.}
  \label{fig:realintrinsic}
\end{figure}

\paragraph{Generalization Capabilities}
Teamwork's ability to leverage pretrained diffusion models can aid in
generalizing beyond the training data.  To evaluate Teamwork's
generalization capabilities, we compare
in~\autoref{tab:generalization} the InteriorVerse trained Intrinsic
Image Diffusion~\cite{Kocsis:2023:IID},
RGB$\rightarrow$X~\cite{Zeng:2024:RGB}, and the intrinsic image
decomposition Teamwork model on (common intrinsic components from) the
HyperSim~\cite{Roberts:2021:HPS} and
Infinigen~\cite{Raistrick:2023:IPW} test sets. Note that existing
real-world intrinsic datasets~\cite{Wu:2023:MAI} only assume diffuse
albedo and are therefore not suited for validating models with an
expanded set of intrinsic components. While HyperSim also features
indoor scenes, Infinigen contains outdoor scenes, and thus is more
distinct from the InteriorVerse training data. Over both test sets,
our Teamwork variant exhibits better generalization capabilities than
competing diffusion-based intrinsic decomposition methods.
\autoref{fig:realintrinsic} further qualitatively demonstrates
Teamwork's generalization capabilities on real-world photographs. We
observe that Intrinsic Image Diffusion sometimes fails to extract
shadows and reflections, while RGB$\rightarrow$X struggles to produce
consistent normals. Refer to the supplemental material for more
real-world side-by-side comparisons.

\section{Conclusion}
\label{sec:conclusion}
In this paper, we presented Teamwork, an efficient and flexible
unified framework for adapting and expanding the number of input and
output channels of a pretrained large image diffusion model. Teamwork
leverages a novel variation of LoRA to coordinate and adapt between
multiple instances of a base diffusion model.  We introduced a novel
way to add additional control signals to a model as well as an easy
method of dynamically activating different teammates.  We demonstrated
that Teamwork performs similarly or better than prior work that relies
on bespoke solutions for a variety of diffusion-derived graphics tasks
such as inpainting, SVBRDF estimation, intrinsic decomposition, neural
shading, and intrinsic image synthesis.

\begin{acks}
This research was supported in part by NSF grant IIS-1909028.
\end{acks}

\bibliographystyle{ACM-Reference-Format}
\bibliography{references}


\end{document}